\documentclass[sigconf,nonacm]{acmart}

\AtBeginDocument{%
  }

\setcopyright{none}
\acmConference[]{}{}{}
\acmBooktitle{}
\renewcommand\footnotetextcopyrightpermission[1]{}
\usepackage[utf8]{inputenc} % allow utf-8 input
\usepackage[T1]{fontenc}    % use 8-bit T1 fonts
\usepackage{hyperref}       % hyperlinks
\usepackage{url}            % simple URL typesetting
\usepackage{booktabs}       % professional-quality tables
\usepackage{amsfonts}       % blackboard math symbols
\usepackage{nicefrac}       % compact symbols for 1/2, etc.
\usepackage{microtype}      % microtypography
\usepackage{xcolor}         % colors
\usepackage{wrapfig}
\usepackage{enumitem}
\usepackage{tikz}
\usepackage{algorithm}
\usepackage{algpseudocode}

\usepackage[font=small,skip=2pt]{caption} 
\usepackage{subcaption}

\usepackage{adjustbox} % for robust sizing

\usepackage{cleveref}

\definecolor{mydarkblue}{rgb}{0,0.08,0.45}
\hypersetup{ %
pdftitle={},
pdfsubject={},
pdfkeywords={},
pdfborder=0 0 0,
pdfpagemode=UseNone,
colorlinks=true,
linkcolor=mydarkblue,
citecolor=mydarkblue,
filecolor=mydarkblue,
urlcolor=mydarkblue,
}

\usepackage{algorithm}
\usepackage{algpseudocode}

\usepackage{booktabs}      % nice horizontal rules
\usepackage{multirow}      % multi-row cells
\usepackage{array}         % for p{..}-style columns
\usepackage{tabularx}
\usepackage{makecell}  

\usepackage{tikz}
\usepackage{amsmath}                % for \mathcal etc.
\usetikzlibrary{matrix}             % used for the legend

\usepackage{amsmath}
\usepackage{mathtools}
\usepackage{amsthm}
\usepackage{thmtools}
\usepackage{thm-restate}

\theoremstyle{plain}

\theoremstyle{definition}

\theoremstyle{definition}

\newtheoremstyle{bolditalicstyle}  % Name of the style
  {3pt}                            % Space above
  {3pt}                            % Space below
  {\itshape}                       % Body font (italic)
  {}                               % Indent amount
  {\bfseries}                      % Theorem head font (bold)
  {.}                              % Punctuation after theorem head
  {.5em}                           % Space after theorem head
  {}                               % Theorem head spec

\theoremstyle{bolditalicstyle}

\DeclareMathOperator*{\argmin}{argmin}

\begin{document}

\title{Cold-Start Active Correlation Clustering}

% \author{Linus Aronsson}
% \authornote{These authors contributed equally to this work.} % ← footnote text
% \affiliation{%
%   \institution{Chalmers University of Technology \& University of Gothenburg}
%   \country{Sweden}}
% \email{linaro@chalmers.se}

% \author{Han Wu}
% \authornotemark[1]                                    % ← re‑uses the same footnote
% \affiliation{%
%   \institution{Chalmers University of Technology \& University of Gothenburg}
%   \country{Sweden}}
% \email{hanwu@student.chalmers.se}

% \author{Morteza Haghir Chehreghani}
% \affiliation{%
%   \institution{Chalmers University of Technology \& University of Gothenburg}
%   \country{Sweden}}
% \email{morteza.chehreghani@chalmers.se}


\author{Linus Aronsson}
\authornote{These authors contributed equally to this work.} % ← footnote text
\affiliation{%
  \institution{Chalmers University of Technology \& University of Gothenburg}
  \country{Sweden}}
\email{linaro@chalmers.se}

\author{Han Wu}
\authornotemark[1]                                    % ← re‑uses the same footnote
\affiliation{%
  \institution{Chalmers University of Technology \& University of Gothenburg}
  \country{Sweden}}
\email{hanwu@student.chalmers.se}

\author{Morteza Haghir Chehreghani}
\affiliation{%
  \institution{Chalmers University of Technology \& University of Gothenburg}
  \country{Sweden}}
\email{morteza.chehreghani@chalmers.se}

\begin{abstract}
We study active correlation clustering where pairwise similarities are not provided upfront and must be queried in a cost-efficient manner through active learning. Specifically, we focus on the cold-start scenario, where no true initial pairwise similarities are available for active learning. To address this challenge, we propose a coverage-aware method that encourages diversity early in the process. We demonstrate the effectiveness of our approach through several synthetic and real-world experiments.
\end{abstract}

%\keywords{Correlation clustering, active learning, cold-start learning, query efficiency}

\maketitle
% hide conference info & titles & authors in the header
\fancyhead[LE,RO]{}
\fancyhead[LO,RE]{}

\section{Introduction}\label{section:introduction}

\emph{Correlation clustering} (CC)~\cite{BansalBC04,DemaineEFI06} clusters objects directly from the respective signed pairwise relations, accommodating both positive and negative similarities. CC has been used in diverse applications, including image segmentation~\cite{KimNKY11}, bioinformatics~\cite{BonchiGU13}, spam filtering~\cite{BonchiGL14}, social network analysis~\cite{2339530.2339735,2956185, linusmortezaneurips}, duplicate detection~\cite{HassanzadehCML09}, co-reference resolution~\cite{McCallumW04}, entity resolution~\cite{GetoorM12}, color naming~\cite{ThielCD19}, and clustering aggregation~\cite{GionisMT07,ChehreghaniC20}. Computing the optimum is NP-hard and APX-hard~\cite{BansalBC04,DemaineEFI06}; consequently, approximation strategies are employed in practice, with local-search variants often offering a favorable balance of quality and efficiency~\cite{ThielCD19,Chehreghani22_shift}.

In many real-world scenarios, the $\binom{N}{2}$ pairwise similarities needed by CC are \emph{not} available upfront. Obtaining them—e.g., from experts, crowd workers, or laboratory experiments—can be expensive and time-consuming~\cite{bressan2020,bonchi2020}. This motivates \emph{active correlation clustering} (active CC), where the aim is to recover a high-quality CC solution while querying only a small fraction of pairs. We adopt the standard setting considered in prior work~\cite{mazumdar2017-2,bressan2020,bonchi2020,Craenendonck2018COBRASIC,silwal2023kwikbucks,mabcc,anonymous,noisyqecc}: (i) the objective is CC; (ii) pairwise similarities are unknown a priori; (iii) the algorithm may query a single (noisy) oracle under a fixed budget $W \ll \binom{N}{2}$; and (iv) feature vectors are not assumed—information about the clustering is obtained solely from queried pairwise relations.

Early research proposed pivot-based algorithms with query-complexity guarantees under noise~\cite{mazumdar2017-2}, adaptive variants of KwikCluster~\cite{bressan2020,bonchi2020}, and bandit-based formulations~\cite{mabcc,noisyqecc}. While theoretically appealing, these approaches either rely on strong assumptions (e.g., known noise rates) or struggle in realistic noisy regimes. A flexible framework that decouples the query strategy from the downstream CC algorithm was later introduced in~\cite{anonymous}, enabling the design of general query strategies and the use of efficient local-search algorithms~\cite{ThielCD19,Chehreghani22_shift}. Building on this framework, recent work introduced \emph{information-theoretic query strategies} \cite{aronsson2024informationtheoreticactivecorrelationclustering} (based on entropy and information gain) tailored to pairwise querying in CC and reported strong empirical improvements over maxmin/maxexp from \citep{anonymous} and other baselines such as a query-efficient pivot-based approach named QECC \cite{bonchi2020}. See Section \ref{section:experiments} for all baselines.  

Despite their strengths, uncertainty-based methods (e.g., information-theoretic approaches) face two key limitations. (i) They perform poorly in the \emph{cold-start} setting, when no pairwise similarities are initially available. This is because they rely on uncertainty estimates based on the information available so far. This can induce early \emph{selection bias}, where the algorithm repeatedly samples locally informative pairs from a narrow region of the similarity graph before having explored enough of the entire graph. Consequently, many queries may be needed before enough global structure is revealed for the CC algorithm to recover the true clustering. (ii) In batch selection, they often choose pairs that are highly redundant within the same batch, a well-known issue in batch active learning \citep{deep-al-survey, JarlARC22, SAMOAA2024109383, 10386685}.

We address these challenges by proposing a \emph{coverage-aware} query strategy for active CC that explicitly encourages diversity among queried pairs. Intuitively, the method prioritizes broad coverage by querying pairs that span many distinct objects. Our contributions are the following.

%This leads to two key benefitsThis yields an initial representation of the overall similarity structure, which helps to (i) reduce selection bias, and (ii) improve batch diversity, which provides a more reliable foundation for subsequent purely uncertainty-driven criteria. Our approach outperforms existing methods on synthetic and real-world datasets in the cold-start setting (see Section \ref{section:experiments} for our experiments). 

%We evaluate the approach on synthetic and real datasets, emphasizing early-round behavior (cold-start robustness) as well as final clustering quality.% The approach is plug-in and model-agnostic: it fits within the generic active CC framework~\cite{anonymous}, is compatible with noisy oracles, and can be paired with any downstream CC algorithm.

%Our focus is complementary to existing uncertainty-based methods. Rather than proposing a pure uncertainty-based acquisition function, we target the \emph{initialization problem}—how to gather the first valuable signals when nothing is known. The proposed coverage-aware mechanism is simple to implement, adds negligible overhead, and can be used on its own or in combination with existing acquisition functions. We evaluate the approach on synthetic and real datasets under a fixed query budget, emphasizing early-round behavior (cold-start robustness) as well as final clustering quality.

\begin{itemize}[leftmargin=*]
\item We identify and empirically characterize the \emph{cold-start sensitivity} of uncertainty-based query strategies in active CC, linking early-round failures to selection bias and insufficient coverage.
\item We propose a simple and efficient coverage-aware method that prioritizes diversity in queried pairs. This approach offers two key advantages: (i) it promotes diversity within the batch of pairs selected in the current round, thereby mitigating the well-known problem of batch redundancy in batch active learning \citep{deep-al-survey}; and (ii) it promotes diversity between the pairs selected in the current round and those chosen in previous rounds, reducing selection bias and accelerating the accumulation of globally useful information.
\item We demonstrate effectiveness and robustness on synthetic and real datasets, showing consistent gains in the cold-start setting.
\end{itemize}

%Overall, this short paper contributes a practical remedy to the cold-start challenge in active CC. By encouraging coverage in early rounds, the method complements information-theoretic acquisition functions~\cite{aronsson2024informationtheoreticactivecorrelationclustering} and is easy to adopt within existing pipelines that decouple acquisition from clustering~\cite{anonymous}.

\section{Active Correlation Clustering}\label{section:acc-form}

In this section, we formalize active correlation clustering.

\subsection{Problem Setup}

Let $\mathcal{V}=\{1,\dots,N\}$ be the set of vertices (objects) and
$\mathcal{E}=\{(u,v)\mid u,v\in\mathcal{V},\,u<v\}$ the set of (undirected) edges.
We consider a signed, weighted graph
$G=(\mathcal{V},\mathcal{E},\mathbf{S})$, where $\mathbf{S}\in\mathbb{R}^{N\times N}$ is
symmetric with zeros on the diagonal and entries $S_{uv}\in[-1,1]$ serving as \emph{edge
weights}: $+1$ indicates strong similarity, $-1$ strong dissimilarity, and values near
$0$ indicate uncertainty (including oracle ambiguity). Conceptually, CC operates on the
complete signed graph; in the active setting only a small subset of weights is revealed by
querying an oracle. We maintain an estimate $\mathbf{S}$ of the unknown ground-truth
matrix $\mathbf{S}^\ast$, updating entries as queries are answered.

A clustering is a partition of $\mathcal{V}$. We encode a clustering with $K$ clusters as
$\mathbf{c}\in[K]^N$, where $c_u$ is the label of object $u$. We say a pair $(u, v)$ \emph{violates} a clustering $\mathbf{c}$ if $c_u = c_v$ and $S_{uv} < 0$ or $c_u \neq c_v$ and $S_{uv} \geq 0$. The CC objective penalizes violations and can be defined as $R^{\text{CC}}(\mathbf{c}\mid\mathbf{S}) = \sum_{(u,v) \in \mathcal{E}}|S_{uv}|\mathbb{I}[(u, v) \text{ violates } \mathbf{c}]$. This is equivalent, up to an additive constant independent of $\mathbf{c}$, to the \emph{max-correlation} form~\cite{ethz-a-010077098,Chehreghani22_shift}: $R^{\text{MC}}(\mathbf{c}\mid\mathbf{S}) = - \sum_{(u,v)\in\mathcal{E}:~c_u=c_v} S_{uv}$. We have $\argmin_{\mathbf{c}} R^{\text{CC}}(\mathbf{c}\mid\mathbf{S})
= \argmin_{\mathbf{c}} R^{\text{MC}}(\mathbf{c}\mid\mathbf{S})$. We therefore optimize
$R^{\text{MC}}$ (as it leads to a number of simplifications in the derived algorithms). The ground-truth clustering is
$\mathbf{c}^\ast=\argmin_{\mathbf{c}} R^{\text{MC}}(\mathbf{c}\mid\mathbf{S}^\ast)$.
\subsection{Active CC Procedure}
\begin{algorithm}[tb]
\caption{Generic Active CC}
\label{alg:acc}
\begin{algorithmic}[1]
\Require initial weights $\mathbf{S}^{0}$, batch size $B$, total query budget $W$, query strategy $\mathcal{S}$
\State $i \gets 0$, \ $q \gets 0$
\While{$q < W$}
  \State $\mathbf{c}^{i} \gets \textsc{CC-Algorithm}(\mathbf{S}^{i})$
  \State Select a batch $\mathcal{B} = \mathcal{S}(\mathbf{S}^i, \mathbf{c}^i)\subseteq\mathcal{E}$ of size $B$
  \State Query the oracle for all $(u,v)\in\mathcal{B}$ and update the corresponding weights in $\mathbf{S}^{i+1}$
  \State $q \gets q + |\mathcal{B}|$;\quad $i \gets i + 1$
\EndWhile
\State \Return $\mathbf{c}^{i}$
\end{algorithmic}
\end{algorithm}
We adopt the active CC procedure from \cite{anonymous}, that decouples which edges to query from the downstream CC
algorithm (see Alg. \ref{alg:acc}). At each round, we (i) clusters the current signed graph defined by $\mathbf{S}^{i}$ using any CC  algorithm. We use the local-search CC algorithm from \citep{anonymous}, due to its strong empirical performance. It is highly robust to noise/inconsistency in the similarities, and it dynamically discovers the number of clusters, (ii) selects a batch of edges $\mathcal{B}$ via a query strategy $\mathcal{S}$. Active CC thus comes down to desining effective query strategies. It is common to define $\mathcal{S}$ in terms of an acquisition
function $a:\mathcal{E} \rightarrow \mathbb{R}^+$, where a larger value of $a(u,v)$ indicates greater informativeness of the pair $(u,v)$. The batch $\mathcal{B}$ is then selected by selecting the top-$B$ pairs according to $a$, and (iii) queries the oracle to refine the edge weights in $\mathbf{S}$, based on the selected batch $\mathcal{B}$. The process stops when the query budget $W$ is exhausted. In the cold-start setting, $\mathbf{S}^{0}$ may be uninformative (e.g., all zeros); the coverage-aware choice of $\mathcal{S}$ proposed in this paper is designed to be robust in this setting.

%\begin{algorithm}[tb]
%\caption{Local Search for CC}
%\label{alg:cc-local-search}
%\begin{algorithmic}[1]
%\State Randomly assign each object $u\in\mathcal{V}$ to a cluster
%\While{not converged}
%  \State Select object $i\in\mathcal{V}$ at random
%  \State Move $i$ to the cluster that maximizes the improvement of \eqref{eq:maxcorrfn}; if no existing cluster improves the objective, create a new singleton cluster for $i$
%  \Comment{As in~\cite{anonymous}, evaluating only terms of \eqref{eq:maxcorrfn} involving $i$ reduces per-iteration cost from $O(K^2N^2)$ to $O(KN)$}
%\EndWhile
%\State \Return current clustering $\mathbf{c}$
%\end{algorithmic}
%\end{algorithm}

\subsection{Information-Theoretic Methods}
\label{section:information-theoretic}

We briefly recap the information-theoretic query strategies used in active CC, following recent work on pairwise querying for CC~\cite{aronsson2024informationtheoreticactivecorrelationclustering}. Let $\mathcal{C}$ denote the set of all partitions of $\mathcal{V}$. We define the Gibbs distribution over clusterings with concentration $\beta>0$ as $P^{\text{Gibbs}}(\mathbf{y} = \mathbf{c}) = \exp( -\beta\,R^{\text{MC}}(\mathbf{c}\mid\mathbf{S}))/Z$, where $Z = \sum_{\mathbf{c}'\in\mathcal{C}} \exp(-\beta\,R^{\text{MC}}(\mathbf{c}'\mid\mathbf{S}))$ and $\mathbf{y} \in \mathcal{C}$ is a random vector with sample space $\mathcal{C}$. Direct computation is intractable due to the enumeration of all clustering solutions in $Z$. We approximate $P^{\text{Gibbs}}$ with a factorial distribution
$Q(\mathbf{y})=\prod_{u\in\mathcal{V}} Q(y_u)$, represented by
$\mathbf{Q}\in[0,1]^{N\times K}$ with $Q_{uk}=Q(y_u=k)$. Using variational mean-field~\cite{DBLP:journals/pami/HofmannPB98,pmlr-v22-haghir12},
we alternate the synchronous updates $\mathbf{Q} = \mathrm{softmax}(-\beta\,\mathbf{M})$, and $
\mathbf{M} = -\,\mathbf{S}\,\mathbf{Q}$ until convergence, where $\mathbf{M}\in\mathbb{R}^{N\times K}$ is a matrix of assignment costs (i.e., element $M_{uk}$ should be interpreted as the cost of assigning object $u$ to cluster $k$). The matrix $\mathbf{M}$ can be initialized randomly. In short, this procedure converges to a local minimum of the KL-divergence between $\mathbf{Q}$ and $P^{\text{Gibbs}}$. We refer to \cite{aronsson2024informationtheoreticactivecorrelationclustering} for a detailed description.

\paragraph{Entropy acquisition function.}
Let $E_{uv}\in\{0,1\}$ be a random variable that indicates whether $u$ and $v$ are in the same cluster or not. The same-cluster probability is $P(E_{uv}=1)
\;\approx\;\sum_{k=1}^{K} Q_{uk}\,Q_{vk}$. The \emph{entropy} acquisition function is defined as the entropy of $E_{uv}$ \citep{aronsson2024informationtheoreticactivecorrelationclustering}:
\begin{equation}
\label{eq:entropy}
a^{\text{Entropy}}(u,v)
\coloneqq H(E_{uv}) = \mathbb{E}_{P(E_{uv})}[-\log P(E_{uv})].
\end{equation}
In this paper, we compare against $a^{\text{Entropy}}$ to illustrate the issue of selection bias in uncertainty-based query strategies. We do not include acquisition functions based on expected information gain proposed by \citep{aronsson2024informationtheoreticactivecorrelationclustering}, for three main reasons: (i) they are also subject to selection bias---often more severely than entropy, (ii) their empirical performance is typically similar to entropy, and (iii) they are generally more computationally demanding in practice.

\section{Coverage-Based Query Strategy}
\label{section:coverage}

\begin{figure*}[!t]
  \centering
  \begin{subfigure}[t]{0.25\textwidth}
    \centering
    \includegraphics[width=\linewidth]{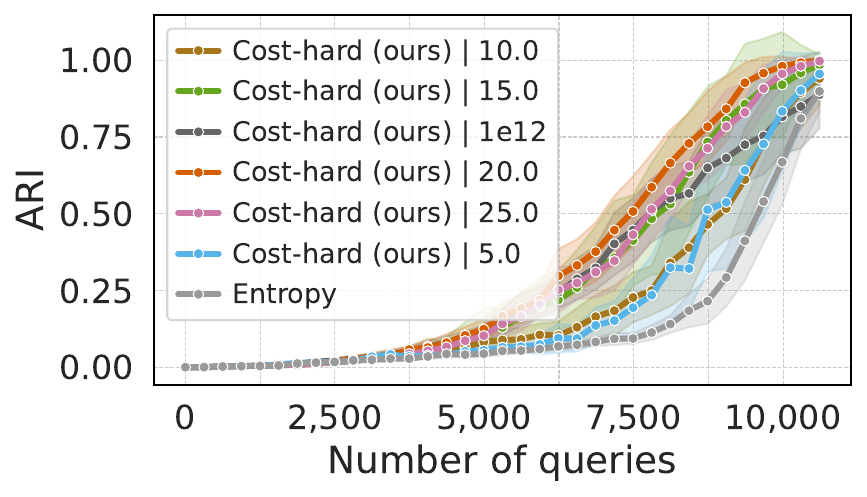}
    \caption{Switch point ablation.}\label{fig:ablA1}
  \end{subfigure}\hfill
  \begin{subfigure}[t]{0.25\textwidth}
    \centering
    \includegraphics[width=\linewidth]{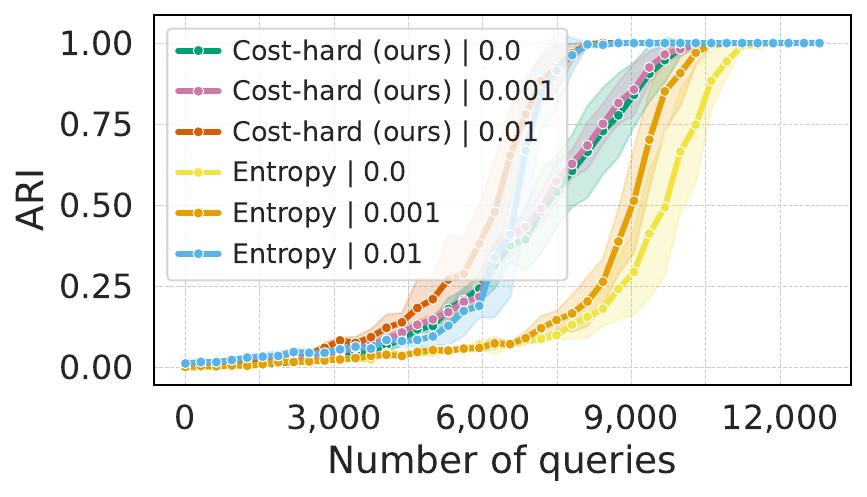}
    \caption{Warm start ablation.}\label{fig:ablA2}
  \end{subfigure}\hfill
  \begin{subfigure}[t]{0.25\textwidth}
    \centering
    \includegraphics[width=\linewidth]{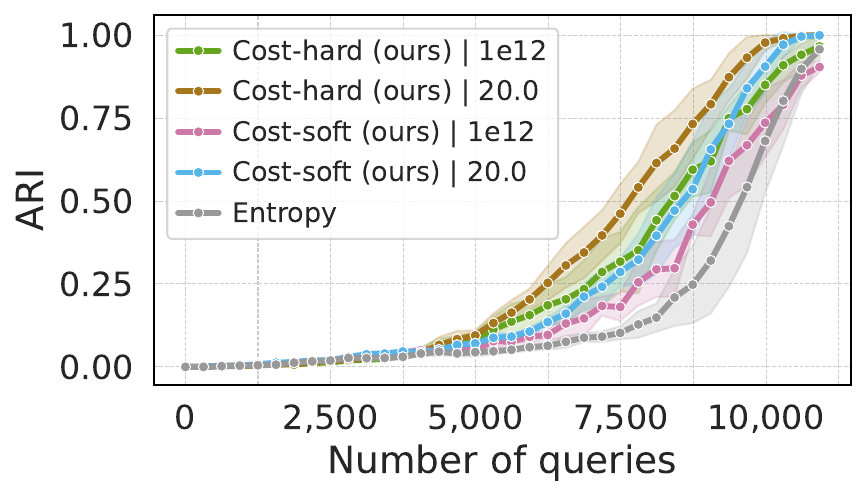}
    \caption{Initialization: Zero}\label{fig:ablC1}
  \end{subfigure}\hfill
  \begin{subfigure}[t]{0.25\textwidth}
    \centering
    \includegraphics[width=\linewidth]{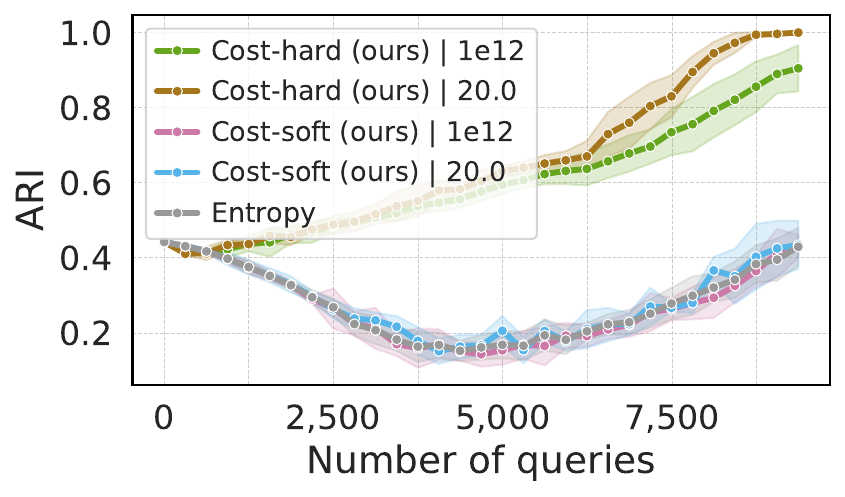}
    \caption{Initialization: KMeans}\label{fig:ablC2}
  \end{subfigure}

  \caption{Ablation studies on the synthetic dataset. See Section \ref{section:experiments} for a detailed description.}
  \label{fig:ablation_all}
\end{figure*}

\begin{figure}[!b]
    \centering
    \begin{subfigure}[!b]{0.5\linewidth}
        \centering
        \includegraphics[width=\linewidth]{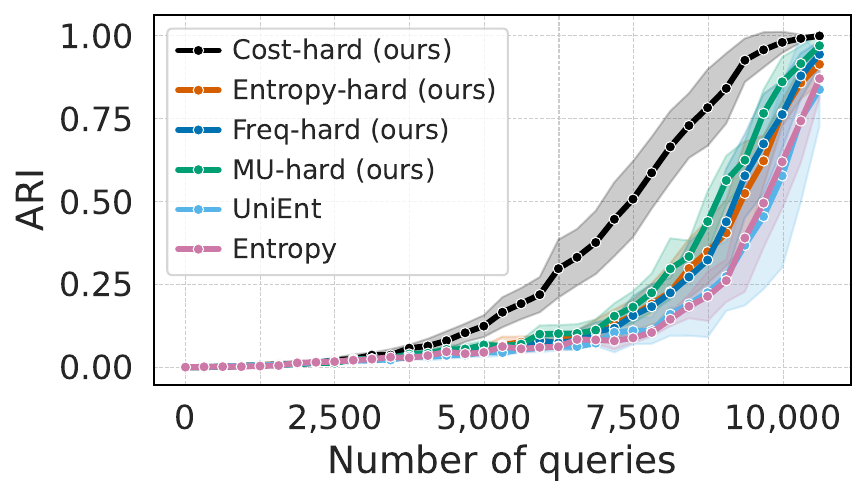}
        \caption{Initialization: Zero}
        \label{fig:diverse_methods_ablation}
    \end{subfigure}\hfill
    \begin{subfigure}[!b]{0.5\linewidth}
        \centering
        \includegraphics[width=\linewidth]{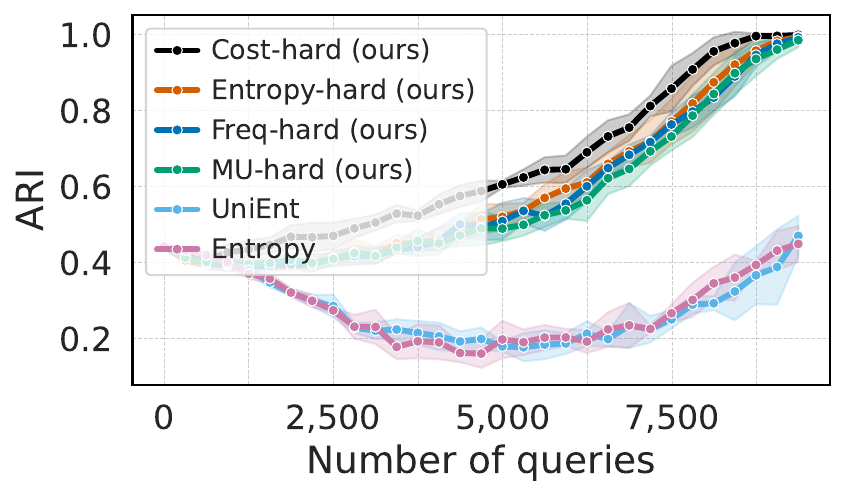}
        \caption{Initialization: KMeans}
        \label{fig:diverse_methods_ablation2}
    \end{subfigure}
    \caption{Comparison of diverse methods on synthetic dataset.}
    \label{fig:diverse_methods}
\end{figure}

\begin{figure*}[!t]
  \centering
  % Row 1
  \begin{subfigure}[t]{0.25\textwidth}
    \includegraphics[width=\linewidth]{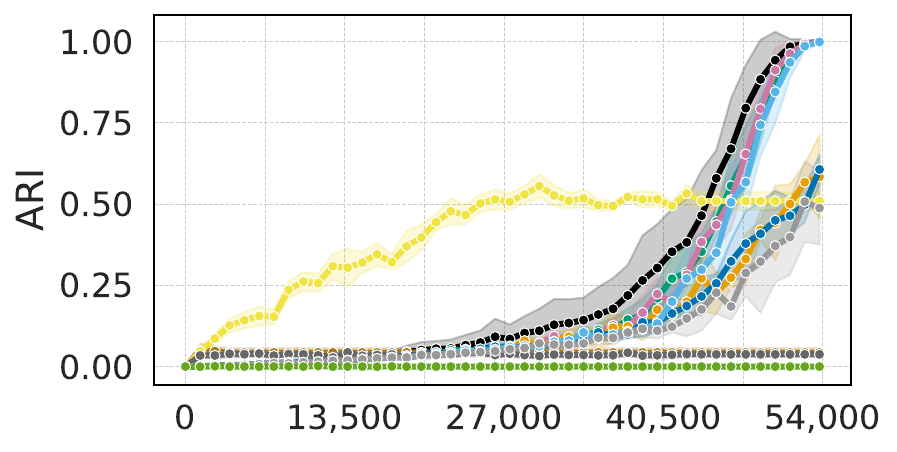}
    \subcaption{20 Newsgroups | Zero}\label{fig:main1}
  \end{subfigure}\hfill
  \begin{subfigure}[t]{0.25\textwidth}
    \includegraphics[width=\linewidth]{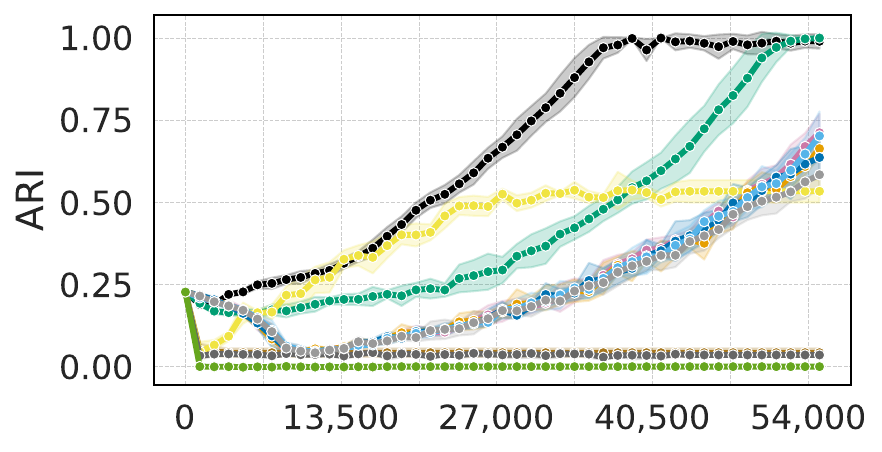}
    \subcaption{20 Newsgroups | KMeans}\label{fig:main2}
  \end{subfigure}\hfill
  \begin{subfigure}[t]{0.25\textwidth}
    \includegraphics[width=\linewidth]{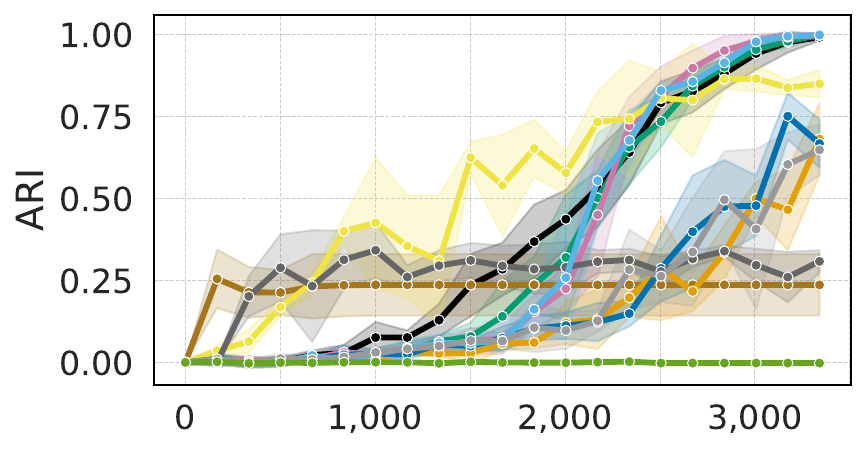}
    \subcaption{ForestTypeMapping | Zero}\label{fig:main3}
  \end{subfigure}\hfill
  \begin{subfigure}[t]{0.25\textwidth}
    \includegraphics[width=\linewidth]{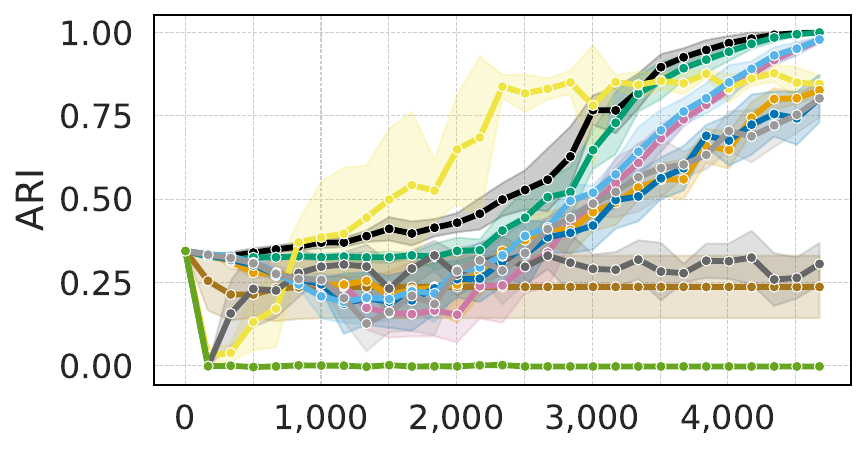}
    \subcaption{ForestTypeMapping | KMeans}\label{fig:main4}
  \end{subfigure}

  \vspace{-1pt} % reduce gap between rows

  % Row 2
  \begin{subfigure}[t]{0.25\textwidth}
    \includegraphics[width=\linewidth]{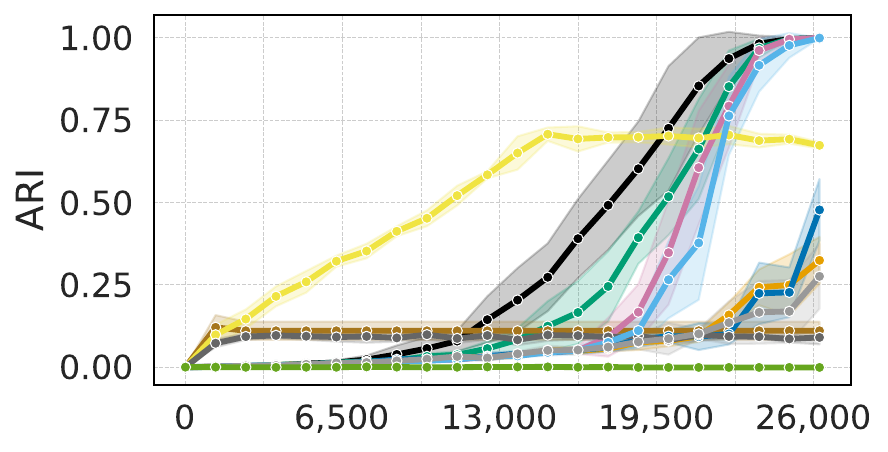}
    \subcaption{Cifar10 | Zero}\label{fig:main5}
  \end{subfigure}\hfill
  \begin{subfigure}[t]{0.25\textwidth}
    \includegraphics[width=\linewidth]{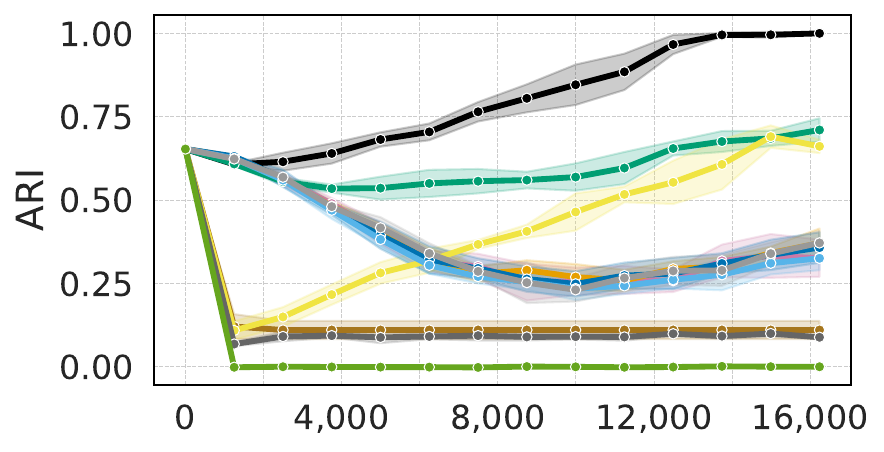}
    \subcaption{Cifar10 | KMeans}\label{fig:main6}
  \end{subfigure}\hfill
  \begin{subfigure}[t]{0.25\textwidth}
    \includegraphics[width=\linewidth]{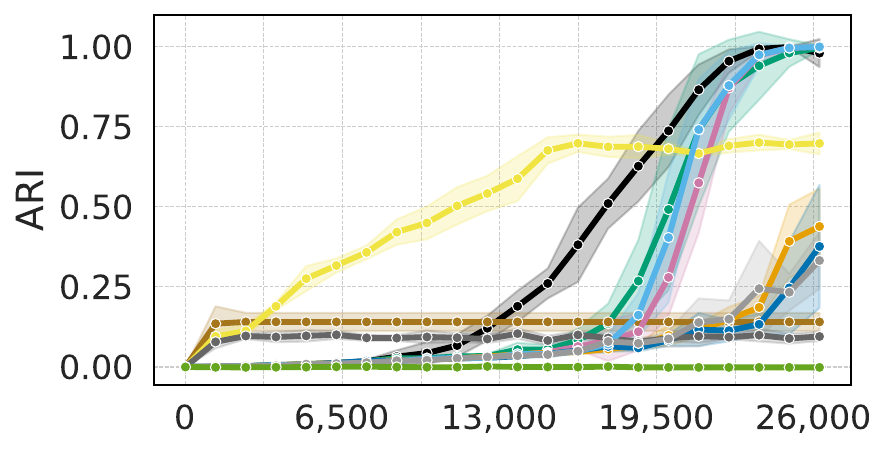}
    \subcaption{MNIST | Zero}\label{fig:main7}
  \end{subfigure}\hfill
  \begin{subfigure}[t]{0.25\textwidth}
    \includegraphics[width=\linewidth]{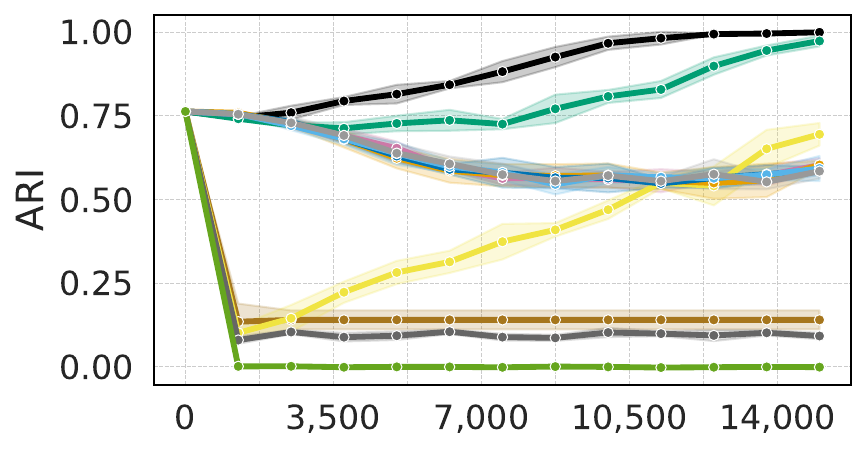}
    \subcaption{MNIST | KMeans}\label{fig:main8}
  \end{subfigure}

  \vspace{-1pt} % reduce gap between rows

  % Row 3
  \begin{subfigure}[t]{0.25\textwidth}
    \includegraphics[width=\linewidth]{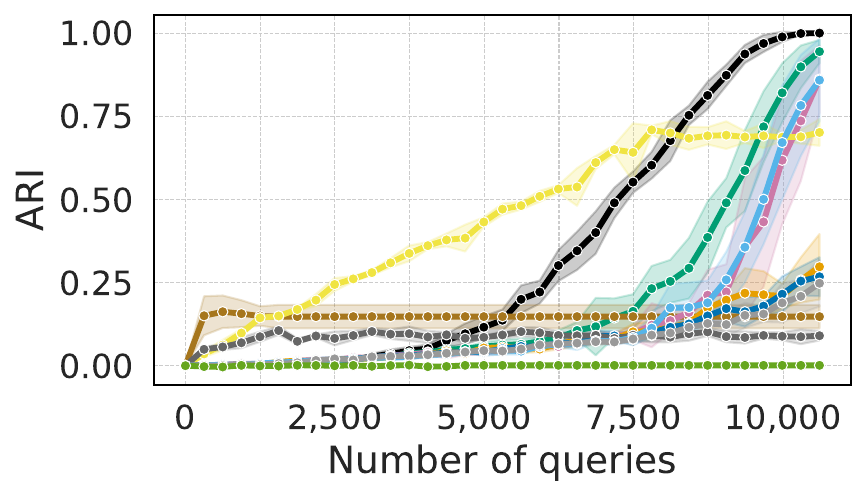}
    \subcaption{Synthetic | Zero}\label{fig:main9}
  \end{subfigure}\hfill
  \begin{subfigure}[t]{0.25\textwidth}
    \includegraphics[width=\linewidth]{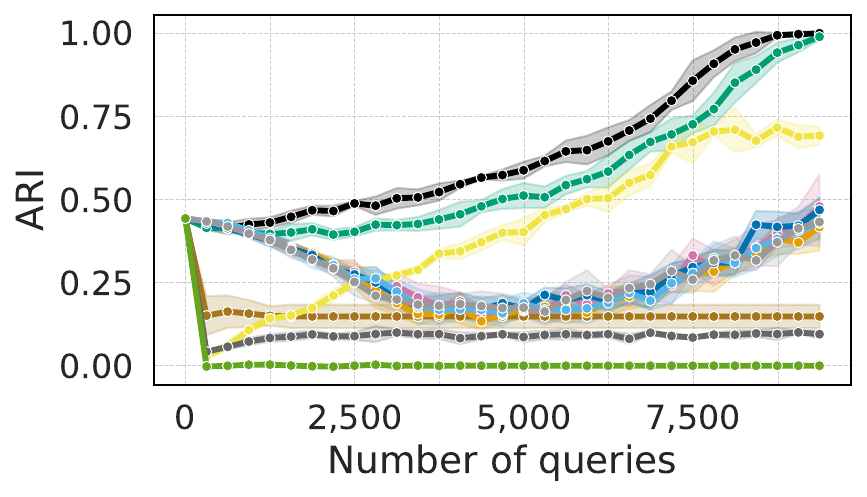}
    \subcaption{Synthetic | KMeans}\label{fig:main10}
  \end{subfigure}\hfill
  \begin{subfigure}[t]{0.25\textwidth}
    \includegraphics[width=\linewidth]{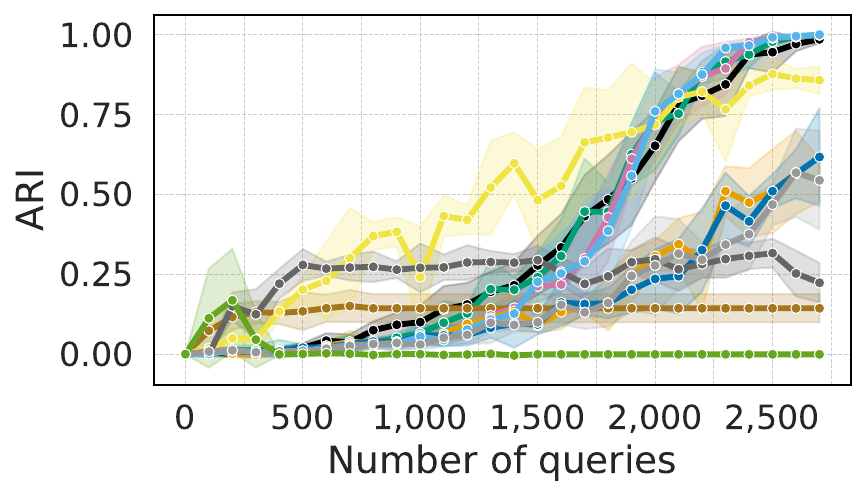}
    \subcaption{User Knowledge | Zero}\label{fig:main11}
  \end{subfigure}\hfill
  \begin{subfigure}[t]{0.25\textwidth}
    \includegraphics[width=\linewidth]{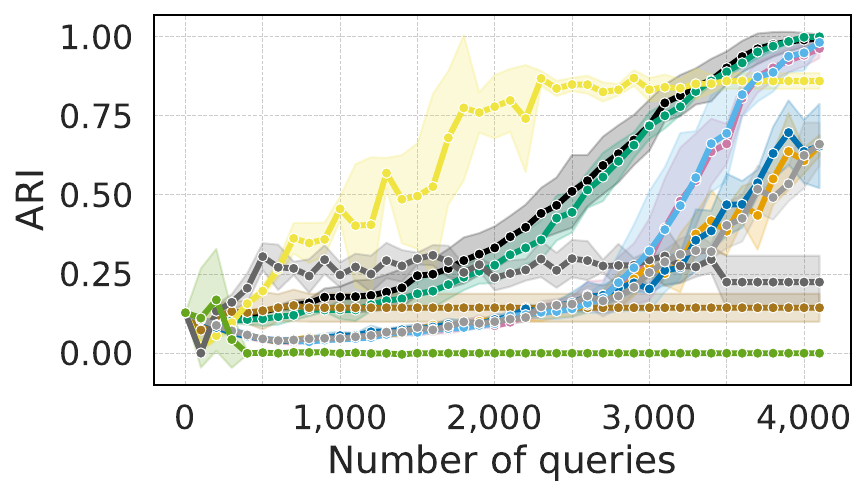}
    \subcaption{User Knowledge | KMeans}\label{fig:main12}
  \end{subfigure}

  \vspace{-1pt} % reduce space before legend

  \includegraphics[width=.7\textwidth]{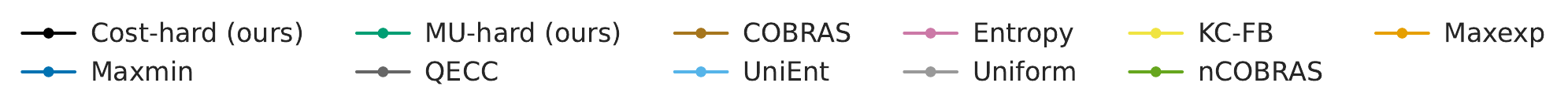}

  \caption{Results for different methods across datasets.}
  \label{fig:main}
\end{figure*}

To deal with cold-start selection bias (and batch redundancy), we propose to group edges into \emph{query regions} and
allocate the batch budget $B$ across regions in proportion to their \emph{size-normalized} informativeness. We allow either \emph{soft} region memberships (using the mean-field matrix $\mathbf{Q}$) or \emph{hard} memberships (from the current clustering $\mathbf{c}^i$). We present the methods with arbitrary matrix $\mathbf{U} \in [0,1]^{N\times K}$, which covers both the soft and hard case (since we can construct a hard variant of $\mathbf{U}$ by setting $U_{uk} = \mathbb{I}[c^i_u = k]$ for all $u \in \mathcal{V}$ and $k \in [K]$). 

\paragraph{Definition of query regions.} The set of query regions is a partition of the pairs $\mathcal{E}$. While the regions could be defined in many different ways, we propose to construct them given the current clustering solution $\mathbf{c}^i \in \mathcal{C}$ with $K$ clusters. We use $\mathcal{R}=\{(a,a)\}_{a=1}^K \cup \{(a,b)\}_{1\le a<b\le K}$ to represent the query regions. We then use $R_{(a,a)}=\{(u,v):c^i_u=c^i_v=a\}$ and
$R_{(a,b)}=\{(u,v):\{c^i_u,c^i_v\}=\{a,b\}\}$ for $a<b$ to denote the pairs in each region. This means that each region is either all pairs inside a cluster $a \in [K]$, or all pairs going between any two clusters (when $a < b)$. Notably, the number of clusters $K$ can vary between iterations, since the CC algorithm used dynamically determines the number of clusters given the similarities queried so far. The regions in $\mathcal{R}$ is thus adaptive to the iteration $i$ of Alg. \ref{alg:acc} both in terms of (i) which objects belong to each cluster, and (ii) the total number of clusters $K$. 
\paragraph{Query region sizes.} For any edge $(u,v)$ and cluster indices $a,b\in\{1,\ldots,K\}$, we define the region membership weights 
\begin{equation}
w^{(a,a)}_{uv}=U_{ua}U_{va}, \quad w^{(a,b)}_{uv}=U_{ua}U_{vb}+U_{ub}U_{va} \text{ for } a<b.
\end{equation}
Let $s =\mathbf{U}^\top \mathbf{1}_N \in \mathbb{R}^{K}$ (each element is then $s_a=\sum_{u} U_{ua}$) and $\mathbf{B}=\mathbf{U}^\top\mathbf{U}$. The (soft) number
of edges attributable to each region is $N_{aa} =\sum_{u<v} w^{(a,a)}_{uv} = \tfrac{1}{2}(s_a^2 - B_{aa})$ and $N_{ab}=\sum_{u<v} w^{(a,b)}_{uv} = s_a s_b - B_{ab}$ for $(a<b)$. If $\mathbf{U}$ represent a hard assignment, i.e., $U_{ua}=\mathbb{I}\{c^i_u=a\}$, then $N_{aa}=|R_{(a,a)}|$ and $N_{ab}=|R_{(a,b)}|$ for $(a<b)$. Thus, the region sizes reduce to the usual counts of within- and between-cluster pairs.

\paragraph{Region informativeness mass.}
Let $\mathbf{A}\in\mathbb{R}_{\ge 0}^{N\times N}$ be a symmetric matrix, with $A_{uu}=0$, where each element $A_{uv}$ represents some notion of informativeness of the pair $(u, v)$. The total (soft) \emph{informativeness mass} in each region is $M_{aa}=\sum_{u<v} w^{(a,a)}_{uv}A_{uv}=\tfrac{1}{2}G_{aa}$ and $M_{ab}=\sum_{u<v} w^{(a,b)}_{uv}\,A_{uv}=G_{ab}$ for $a<b$ where $\mathbf{G}=\mathbf{U}^\top \mathbf{A}\mathbf{U} \in \mathbb{R}^{K \times K}$. We use the vectorized
forms via $\mathbf{G}$ in practice for efficiency. The purpose of defining a per-region value mass using an arbitrary matrix $\mathbf{A}$ is to establish a flexible framework in which queries can be distributed across regions in any manner, thereby enabling a fully general and adaptable setup. 

\paragraph{Region informativeness normalized by region size.}
We normalize by region size to avoid bias toward large regions to obtain the final score $V_r=M_r/\max(N_r,\varepsilon)$ for each region $r \in \mathcal{R}$ ($\varepsilon>0$ is used for stability). Then, the proportion of queries $\pi_r \in [0,1]$ (with $\sum_r \pi_r = 1)$ to be made in region $r \in \mathcal{R}$ is computed as in Eq. \eqref{eq:region-mixture}.
\begin{equation} 
\label{eq:region-mixture}
\pi_r=\frac{V_r}{\sum_{s\in\mathcal{R}} V_s},
\end{equation}
%
%\begin{algorithm}[tb]
%\caption{Compute Coverage-Aware Region Proportions $\{\pi_r\}$}
%\label{alg:coverage-mixture}
%\begin{algorithmic}[1]
%\Require $\mathbf{Q}\in[0,1]^{N\times K}$ (soft or hard memberships), $\mathbf{A}\in\mathbb{R}_{\ge 0}^{N\times N}$ (symmetric, $A_{uu}=0$), stability $\varepsilon>0$
%\Statex \emph{Cluster indices $a,b\in\{1,\ldots,K\}$ with $a<b$.}
%\State $s \gets \mathbf{Q}^\top \mathbf{1}_N$;\quad $\mathbf{B} \gets \mathbf{Q}^\top \mathbf{Q}$;\quad $\mathbf{G} \gets \mathbf{Q}^\top \mathbf{A}\,\mathbf{Q}$
%\State $N_{aa} \gets \tfrac{1}{2}\big(s_a^2 - B_{aa}\big)$ for all $a$; $N_{ab} \gets s_a s_b - B_{ab}$ for all $a<b$
%\State $M_{aa} \gets \tfrac{1}{2}\,G_{aa}$ for all $a$;\quad $M_{ab} \gets G_{ab}$ for all $a<b$
%\State $V_{aa} \gets M_{aa} / \max(N_{aa},\varepsilon)$ for all $a$;\quad $V_{ab} \gets M_{ab} / \max(N_{ab},\varepsilon)$ for all $a<b$
%\State $Z \gets \sum_{a=1}^{K} V_{aa} \;+\; \sum_{1\le a<b\le K} V_{ab}$
%\State $\pi_{aa} \gets V_{aa}/Z$ for all $a$;\quad $\pi_{ab} \gets V_{ab}/Z$ for all $a<b$
%\State \Return $\{\pi_{aa}\}_{a=1}^K \cup \{\pi_{ab}\}_{1\le a<b\le K}$
%\end{algorithmic}
%\end{algorithm}
\paragraph{Choice of matrix $\mathbf{A}$.}
We instantiate $\mathbf{A}$ in several ways, depending on what we want the region proportions $\{\pi_r\}$ to emphasize. (i) \emph{Entropy:} $A^{\text{Entropy}}_{uv}=a^{\text{Entropy}}(u,v)$ from Eq. \eqref{eq:entropy}, which will prioritize regions with large uncertainty according to the mean-field approximation $\mathbf{Q}$. (ii) \emph{CC-cost contribution:} $A^{\text{Cost}}_{uv}=\lvert S_{uv}\rvert\cdot \mathbb{I}[(u,v)\ \text{violates }\mathbf{c}^i]$ (based on the CC cost $R^{\text{CC}}(\mathbf{c}\mid\mathbf{S})$). This targets edges that are immediately relevant to reducing the CC objective. For example, if a cluster contains many negative edges (i.e., a high CC cost within the cluster), this likely indicates that the cluster should be split into two or more smaller clusters. Such inconsistencies can be resolved by querying additional similarities within the cluster. (iii) \emph{Frequency:} $A^{\text{Freq}}_{uv}=1-F_{uv}$ with $F_{uv}\in\{0,1\}$ indicating whether $(u,v)$ has already been queried. This encourages broad coverage by prioritizing regions with many unqueried pairs relative to the region size. (iv) \emph{Magnitude uncertainty (MU):} $A^{\text{MU}}_{uv}=1-\lvert S_{uv}\rvert$ (recall $S_{uv}\in[-1,1]$), giving higher scores to pairs whose current similarity estimates are near $0$. %These choices provide different ways to quantify the total informativeness of a region.

\paragraph{Batch allocation and within-region selection.}
Given region proportions $\{\pi_r\}$ and batch size $B$, allocate $B_r = \operatorname{round}(\pi_r B)$ queries to each region $r$, using a largest-remainder adjustment so that $\sum_r B_r = B$ and $B_r \geq 0$. For example, given a region $(a, b) \in \mathcal{R}$, we select exactly $B_{(a,b)}$ pairs from the set $R_{(a,b)}$. If $\lvert R_{(r)} \rvert < B_r$, all pairs in the region is queried, and the remaining budget $B_r - \lvert R_{(r)} \rvert$ is allocated to other regions. For any pair $(u, v) \in r$ for some region $r \in \mathcal{R}$, we define the probability of selecting pair $(u,v)$ within region $r$ as $p(u,v \mid r) = a^{\text{Entropy}}(u,v) / \sum_{(w,z) \in R_{(r)}} a^{\text{Entropy}}(w,z)$. From each region, we then sample $B_r$ pairs without replacement according to this distribution. Equivalently, this is the same as selecting the top-$B_r$ pairs with respect to the modified acquisition function $a(u,v) = \log \big(a^{\text{Entropy}}(u,v)\big) + \epsilon_{uv}$ with $\epsilon_{uv} \sim \text{Gumbel}(0,1)$, restricted to pairs $(u,v) \in R_{(r)}$. This approach balances uncertainty-driven selection (via $a^{\text{Entropy}}$) with exploration via sampling. Empirically, it performs substantially better than directly selecting the top-$B_r$ pairs with $a^{\text{Entropy}}$. Combining the methods for computing region proportions (soft or hard) with a given matrix $\mathbf{A}$ (Entropy, Cost, Freq, or MU) yields 8 variants.

\section{Experiments} \label{section:experiments}

In this section, we present our experimental setup and results, closely following the protocol of \citep{aronsson2024informationtheoreticactivecorrelationclustering}. Our evaluation uses one synthetic dataset (with 10 size-balanced clusters) and five real-world datasets: CIFAR-10 \cite{cifar10}, 20 Newsgroups \cite{uci}, Forest Type Mapping \cite{uci}, User Knowledge Modeling \cite{uci}, and MNIST \cite{DBLP:journals/pieee/LeCunBBH98}. Unless otherwise specified, experiments are conducted on the synthetic dataset. For each dataset, we use at most $N = 1000$ data instances, consistent with \citep{aronsson2024informationtheoreticactivecorrelationclustering}, since some baseline methods are computationally expensive (although our methods scale to much larger datasets). Data preprocessing follows \citep{aronsson2024informationtheoreticactivecorrelationclustering}, with the exception that for 20 Newsgroups we construct the dataset using samples from all 20 topics.

In addition, we follow \cite{aronsson2024informationtheoreticactivecorrelationclustering} and adopt the same CC algorithm, noisy oracle, evaluation metric, and baselines. The oracle returns the ground-truth similarity ($+1$ if two instances belong to the same class and $-1$ otherwise) with probability $1-\gamma$, and a random value in $[-1,+1]$ with probability $\gamma$, where we fix $\gamma = 0.4$. At each iteration of the active CC procedure, we compute the adjusted rand index (ARI) between $\mathbf{c}^i$ and the ground-truth clustering (given by the true class labels of each dataset). The baselines include entropy from \citep{aronsson2024informationtheoreticactivecorrelationclustering} (Eq. \eqref{eq:entropy}), where we apply the sampling approach described at the end of Section~\ref{section:coverage} to improve batch diversity, following \citep{aronsson2024informationtheoreticactivecorrelationclustering}; maxmin and maxexp from \cite{anonymous}, which originally introduced the active CC procedure in Alg.~\ref{alg:acc}; a pivot-based active CC algorithm called QECC \citep{bonchi2020}; two adapted state-of-the-art active constraint clustering methods COBRAS \citep{Craenendonck2018COBRASIC} and nCOBRAS \citep{lirias3060956}; and a recent bandit-based approach KC-FB \citep{noisyqecc}. Finally, we include a simple baseline, denoted \emph{UniEnt}, that selects pairs randomly for a few iterations before switching to entropy. After empirical tuning on each dataset, we fix the number of iterations before switching to 20 for the synthetic dataset and 10 for the real-world datasets. This baseline highlights that our approach outperforms naive random exploration, a common strategy for mitigating selection bias. Importantly, we query each pair at most once. 

We consider two strategies for initializing the similarity matrix $\mathbf{S}^0$: (i) all similarities are set to \emph{zero}, representing no prior knowledge; and (ii) we apply $k$-means clustering on the feature vectors of each dataset and set $S^0_{uv} = 0.01$ if $(u,v)$ are assigned to the same cluster and $-0.01$ otherwise. The second approach incorporates weak prior knowledge about the true clustering but may introduce bias if the feature space is noisy, potentially leading to selection bias. Unless otherwise specified, we use the zero initialization.

It is reasonable to assume that once sufficient information about the true similarities has been collected, one can safely \emph{switch} to a purely uncertainty-driven strategy without suffering from selection bias. Our first experiment investigates this hypothesis (Figure~\ref{fig:ablA1}) by evaluating the performance of our method \emph{cost-hard} when switching to entropy at different iterations. For reference, we also include pure entropy (i.e., starting from iteration~0). We find that our method consistently outperforms pure entropy across all switch points, demonstrating robustness to the choice of when to switch. This highlights the potential for future work on dynamically determining the optimal switch point. Empirically, switching after 20 iterations yields the best performance, surpassing even the case of never switching ($1e12$). Based on these findings, we fix the switch point to 20 for the synthetic dataset and 10 for all real-world datasets in the remaining experiments (empirically chosen).

In the next experiment (Figure~\ref{fig:ablA2}), we study the effect of varying degrees of \emph{warm-start}. Specifically, we compare the performance of our method cost-hard and entropy as we vary the proportion of ground-truth similarities revealed at initialization. We find that entropy performs very well when provided with substantial initial information (proportion $0.01$), but degrades significantly under limited initial knowledge ($0$ or $0.001$) due to selection bias, whereas our method remains more robust. Importantly, this experiment assumes access to perfect (noise-free) oracle information, which is unrealistic in practice and underscores the need for methods that perform well in the cold-start regime. Furthermore, note that $0.01\%$ of all pairs in a dataset with $N=5000$ corresponds to about $125000$ pairs known in advance, which is clearly impractical.

In Figures \ref{fig:ablC1}-\ref{fig:ablC2}, we compare the performance of the \emph{soft} and \emph{hard} region membership approaches under two different switch points. Overall, the hard region approach performs better across both initialization strategies. In particular, with $k$-means initialization, the soft approach is clearly affected by selection bias, similar to entropy, likely because it also relies on uncertainty estimates from $\mathbf{Q}$. Consequently, we adopt the hard membership approach in all subsequent experiments. In Figure~\ref{fig:diverse_methods}, we evaluate different choices of $\mathbf{A}$ (cost, entropy, freq, MU). Among these, cost-hard achieves the best overall performance, followed by MU-hard, and we therefore focus on these two methods in the remaining experiments. We also observe that UniEnt is consistently outperformed by all of our methods, indicating that our approaches provide a stronger form of initial exploration than simple random exploration.

%In the main experiment, we adopt a two-stage query strategy for our proposed methods: on the synthetic dataset, we switched to the Entropy strategy after the 20th iteration, and on the remaining real-world datasets, the switch is made after the 10th iteration. We use the entropy information matrix in all experiments. Furthermore, considering the differences in data complexity, when initializing the similarity matrix with KMeans, we set $k=20$ for 20newsgroups, CIFAR10 and MNIST, and $k=10$ for all other datasets. All experiments are conducted on CPU nodes with dual Intel Xeon Platinum 8358 CPUs, each with 64 cores and 512 GB of memory.

% Should we describe the two different initialization strategies in this part?

Finally, Figure~\ref{fig:main} presents the results for all methods across all datasets and both initialization strategies. Overall, we observe that our methods reach $\text{ARI} = 1$ more quickly than the baseline methods on most datasets, demonstrating the effectiveness of our approach in cold-start scenarios.

%We further conduct ablation studies on the synthetic dataset. We first investigate the influence of different switch points on the performance of our Cost-hard method. As shown in \Cref{fig:switch_points_ablation}, the performance of our method largely depends on the choice of switch points, suggesting that its proper selection can be a promising future study direction. Nevertheless, across a wide range of switch points, our method still consistently outperforms the Entropy method. Secondly, we analyze the differences between warm start and cold-start. As shown in \Cref{fig:warm_start_ablation}, a higher proportion of warm start (that is, initializing with more data samples) can significantly improve performance, while for the same warm start proportion, our method always outperforms the Entropy method. We then compare the performance of the four proposed methods. As shown in \Cref{fig:diverse_methods_ablation}, all of our proposed methods generally outperform the Entropy method, with Cost-hard performing the best. Finally, we evaluate the two region modes with different initialization strategies under the switch and non-switch scenarios. As shown in \Cref{fig:soft_region_ablation}, the hard region mode always outperforms the soft region mode and the Entropy method.

\section{Conclusion}

We proposed a coverage-aware query strategy for cold-start active correlation clustering that promotes diversity in the selected pairwise similarities. Experiments on synthetic and real datasets showed that our methods consistently reduce selection bias and discovers the ground-truth clustering faster than existing baselines.

\section*{Acknowledgments}

The work of Linus Aronsson and Morteza Haghir Chehreghani was partially supported by the Wallenberg AI, Autonomous Systems and Software Program (WASP) funded by the Knut and Alice Wallenberg Foundation. Finally, the computations and data handling was enabled by resources provided by the National Academic Infrastructure for Supercomputing in Sweden (NAISS), partially funded by the Swedish Research Council through grant agreement no. 2022-06725.

%\section*{Ethical Considerations}
%
%This work focuses on methodological advances in active correlation clustering. 
%While the research itself does not directly process sensitive personal data, potential applications in domains such as social networks, bioinformatics, or image data raise ethical considerations. 
%Possible negative societal impacts include risks of reinforcing biases present in the data, privacy concerns when clustering sensitive information, and misuse of the technology for surveillance or discriminatory purposes. 
%To mitigate such risks, practitioners should carefully consider the choice of datasets, ensure appropriate anonymization where personal data is involved, and evaluate fairness and robustness of clustering outcomes. 
%As our method is intended as a general-purpose algorithmic contribution, we emphasize the responsibility of downstream users to apply it in ethically sound and socially beneficial contexts. 

\bibliographystyle{ACM-Reference-Format}
\bibliography{references}

\end{document}